\documentclass[10pt,twocolumn,letterpaper]{article}

\usepackage{iccv}
\usepackage{times}
\usepackage{epsfig}
\usepackage{graphicx}
\usepackage{amsmath}
\usepackage{amssymb}
\usepackage{booktabs}
\usepackage{multirow}
\usepackage{makecell}
\usepackage[accsupp]{axessibility}  

\usepackage[breaklinks=true,bookmarks=false]{hyperref}

\usepackage[capitalize]{cleveref}
\crefname{section}{Sec.}{Secs.}
\Crefname{section}{Section}{Sections}
\Crefname{table}{Table}{Tables}
\crefname{table}{Tab.}{Tabs.}
\iccvfinalcopy 

\def\iccvPaperID{8719} 

\ificcvfinal\fi

\begin{document}

\title{LISTER:  Neighbor Decoding for Length-Insensitive Scene Text Recognition}

\author{Changxu Cheng,
Peng Wang,
Cheng Da,
Qi Zheng,
Cong Yao\\
DAMO Academy, Alibaba Group\\
{\tt\small \{ccx0127,wdp0072012,dc.dacheng08,zhengqisjtu,yaocong2010\}@gmail.com}
}

\maketitle
\ificcvfinal\thispagestyle{empty}\fi

\begin{abstract}
The diversity in length constitutes a significant characteristic of text. Due to the long-tail distribution of text lengths, most existing methods for scene text recognition (STR) only work well on short or seen-length text, lacking the capability of recognizing longer text or 
performing length extrapolation. This is a crucial issue, since the lengths of the text to be recognized are usually not given in advance in real-world applications, but it has not been adequately investigated in previous works. Therefore, we propose in this paper a method called \textbf{L}ength-\textbf{I}nsensitive \textbf{S}cene \textbf{TE}xt \textbf{R}ecognizer (LISTER), which remedies the limitation regarding the robustness to various text lengths. Specifically, a Neighbor Decoder is proposed to obtain accurate character attention maps with the assistance of a novel neighbor matrix regardless of the text lengths. Besides, a Feature Enhancement Module is devised to model the long-range dependency with low computation cost, which is able to perform iterations with the neighbor decoder to enhance the feature map progressively. To the best of our knowledge, we are the first to achieve effective length-insensitive scene text recognition. Extensive experiments demonstrate that the proposed LISTER algorithm exhibits obvious superiority on long text recognition and the ability for length extrapolation, while comparing favourably with the previous state-of-the-art methods on standard benchmarks for STR (mainly short text)\footnote{https://github.com/AlibabaResearch/AdvancedLiterateMachinery}.
\end{abstract}

\section{Introduction}
Scene text recognition (STR) is a popular topic in the computer vision community~\cite{Shi2015crnn,Shi2016RobustST,Shi2019ASTERAA,Lyu2018MaskTA,Yu2020TowardsAS,Fang2021ReadLH,Na2022MultimodalTR,Wang2022MultiGranularityPF,Bautista2022SceneTR}, which aims at extracting machine-readable symbols from scene text images.
Recently, a variety of works have pushed forward the recognition performance from the perspective of arbitrary-shaped text~\cite{Shi2019ASTERAA,Li2018ShowAA,Zhong2022SGBANetSG}, combining language models~\cite{Shi2019ASTERAA,Yu2020TowardsAS,Fang2021ReadLH,Na2022MultimodalTR,Wang2011EndtoendST}, etc.
However, sequence length, as a vital characteristic of text, is rarely discussed.
In fact, instances of long text occur frequently in websites, compound words, text lines, codes and multi-lingual scenarios~\cite{Chen2021BenchmarkingCT}.
As revealed in previous studies~\cite{GarciaBordils2022OutofVocabularyCR,Wang2019DecoupledAN}, existing methods cannot handle images with long text very well. We regard it as a key issue worthy of in-depth investigation. Concretely, we ought to systematically analyze the performance of existing methods on text of different lengths and explore an effective way to realize length-insensitive STR.


\begin{figure}[t]
   \begin{center}
      \includegraphics[width=0.95\linewidth]{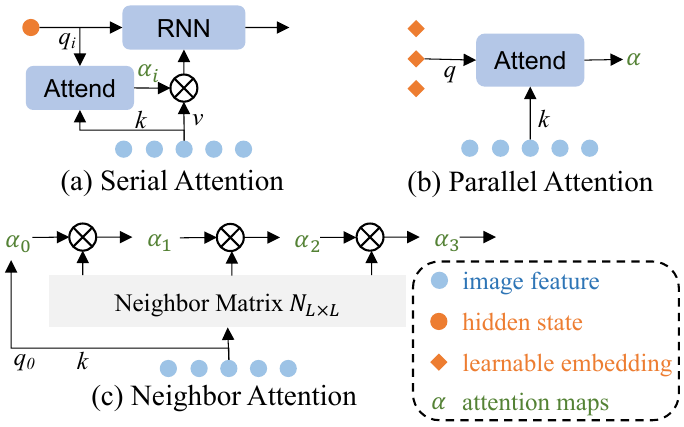}
   \end{center}
   \caption{Different attention mechanisms for STR. $\alpha$ in green is the attention maps produced by: (a) complicated RNN in a serial way (Transformer decoder does similar), (b) pre-defined learnable limited queries in a parallel way, (c) the proposed novel neighbor matrix in a serial but simple and efficient way.}
   \label{intro_att}
   \vspace{-2mm}
\end{figure}

According to our observation, decoders in scene text recognition models are directly associated with the objective of text prediction, thus they are highly likely to play an important role in length-insensitive STR.
In previous works, there are three types of text decoders: CTC~\cite{Graves2006ConnectionistTC,Shi2015crnn}-based decoder, serial attention decoder~\cite{Shi2019ASTERAA,Lu2019MASTERMN,Bautista2022SceneTR}, and parallel attention decoder~\cite{Wang2019DecoupledAN,Yu2020TowardsAS,Fang2021ReadLH}.
The CTC-based decoder makes dense prediction on the feature sequence, and then re-organizes the text by a pre-defined rule. It is efficient, robust to long text recognition, and can be applied to multi-line recognition~\cite{Yousef2020OrigamiNetWS,Wang2021ImplicitFA}, but has some trouble in feature learning~\cite{Liu2018ConnectionistTC,Hu2020GTCGT}.
The serial attention decoder (\cref{intro_att}(a)) adopts RNN or the Transformer decoder to predict characters step by step, which is slow in inference, and may encounter the attention drift problem~\cite{Cheng2017FocusingAT,Wang2019DecoupledAN,Chen2021BenchmarkingCT}, especially for long text.
The parallel attention decoder (\cref{intro_att}(b)) takes pre-defined limited queries to obtain character attention maps in a parallel way, which is efficient as CTC and effective for short text recognition. Thus, it has attracted considerable attention from the community recently.
However, it appears to be sensitive to text length~\cite{GarciaBordils2022OutofVocabularyCR,Wang2019DecoupledAN}.
In principle, each pre-defined query embedding requires a large amount of training data with the corresponding text lengths to learn well. Text of lengths unseen during training are difficult, if not impossible, to be well recognized.

\begin{figure}[t]
   \begin{center}
      \includegraphics[width=\linewidth]{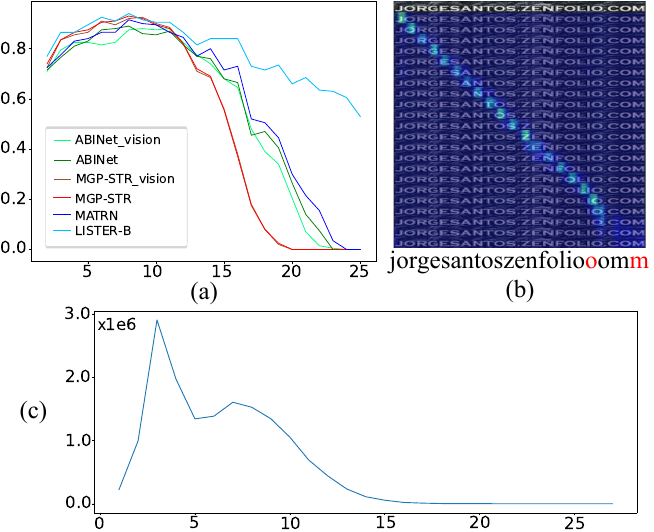}
   \end{center}
   \caption{Performance of recent STR methods on images with different text lengths. (a) Accuracy distribution over length. (b) A case of attention map visualization of ABINet~\cite{Fang2021ReadLH}. (c)Length distribution in the synthetic training set.}
   \label{intro_len}
   \vspace{-3mm}
\end{figure}

To verify this, we simply evaluate the performance of several representative attention-based methods~\cite{Fang2021ReadLH,Wang2022MultiGranularityPF,Na2022MultimodalTR} on a scene text image dataset\footnote{The details of this dataset will be described in ~\cref{sec:datasets}.} where text lengths distribute uniformly.
As shown in \cref{intro_len}(a), the previous methods work very well for short text, but degrade dramatically as the text length becomes larger than 14.
\cref{intro_len}(b) shows the character attention maps where the latter maps are distracted, and the latter characters are also mis-recognized.
Combining with the long-tailed length distribution of the training set~\cite{Jaderberg2014SyntheticDA,Gupta2016SyntheticDF} (\cref{intro_len}(c)), we conjecture that the parallel attention-based methods overfit on the text images of seen lengths seriously.
In brief, the three existing decoders are not able to actualize effective length-insensitive STR.

On the other hand, the long-range feature modeling is crucial to the effectiveness of STR models.
Recently, Transformer layers~\cite{Vaswani2017AttentionIA} have been proved strong for global context modeling~\cite{Yu2020TowardsAS,Fang2021ReadLH,Du2022SVTR,Wang2022MultiGranularityPF,Tan2022PureTW,Bautista2022SceneTR}.
However, there are also concerns about the large computation cost~\cite{Fang2022ABINetAB,Bautista2022SceneTR}.
The computation complexity ($O(N^2)$) increases incontrovertible.
Image patches or token features, including the noisy background, are all fed into the Transformer layers, which is a little bit redundant and expensive for GPU memory~\cite{Zhou2020InformerBE}, especially for a large input size.
Besides, the absolute positional encoding in Transformer layers also restricts the ability for length-insensitive STR.

Aware of the challenges above, we propose a \textbf{L}ength-\textbf{I}nsensitive \textbf{S}cene \textbf{TE}xt \textbf{R}ecognizer (LISTER), which incorporates a novel robust \textit{Neighbor Decoder} (ND) and a lightweight \textit{Feature Enhancement Module} (FEM).
In ND, the attention mechanism is still utilized to ensure the effectiveness.
Like the serial attention-based decoder, we regard the decoding process as a linked list data structure, so each character can be aligned by its previous neighbor regardless of its absolute position in the string.
However, ND relies on a novel neighbor matrix where the next neighbor (character) locations of all the points in the feature map are indicated in a soft way, as shown in \cref{intro_att}(c),
which is simple but effective for length-insensitive text decoding.
To model the long-range feature dependency with low computation cost, we propose to feed only the aligned character features to the Transformer layers and then enhance the whole feature map in FEM.
The self-attention layers adopt the sliding window~\cite{Beltagy2020LongformerTL} to fit arbitrary-length sequences.
Our contributions are summarized as follows:
\begin{itemize}
   \item[1)] A length-insensitive scene text recognizer, \textbf{LISTER}, is proposed with \textit{neighbor decoder} as its core module. To the best of our knowledge, it is the first attempt to realize effective length-insensitive text recognition.
   \item[2)] We propose a \textit{Feature Enhancement Module} that enhance the whole feature map by Transformer layers with low computation cost.
   \item[3)] Through extensive experiments, we prove that LISTER is on par with the previous state-of-the-arts on the common STR benchmarks, while outperforming them on long text. Besides, LISTER also achieves excellent performance in terms of length extrapolation.
\end{itemize}

\begin{figure*}[h]
   \begin{center}
      \includegraphics[width=1.0\linewidth]{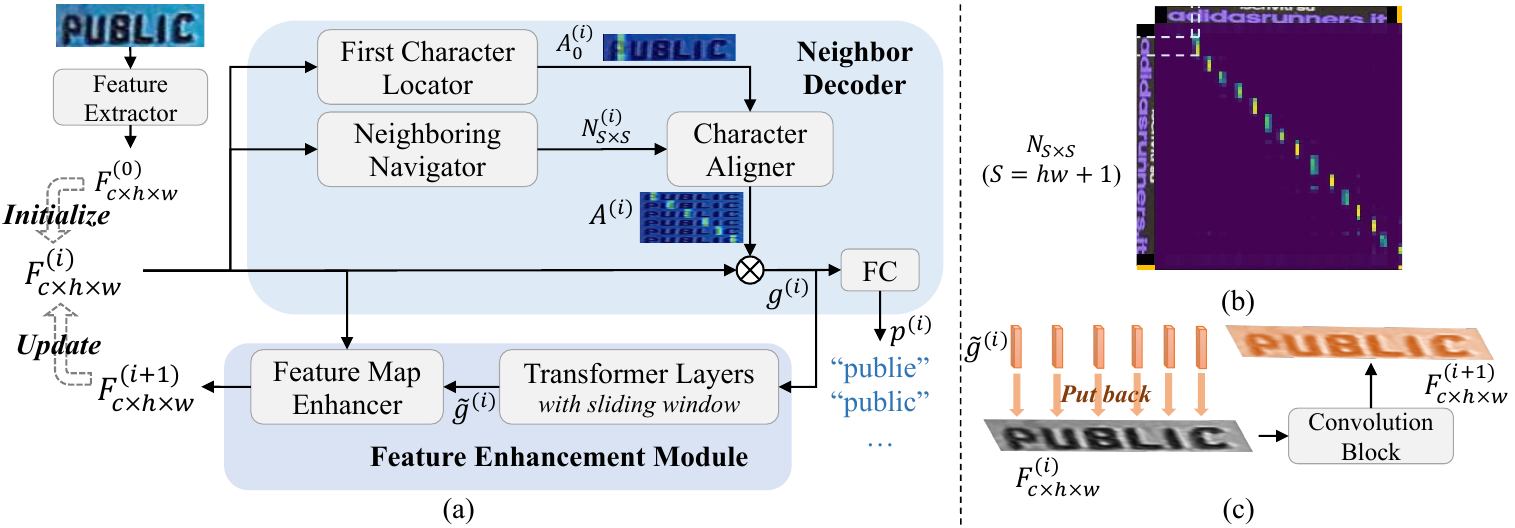}
   \end{center}
   \caption{An overview of LISTER. (a) The model architecture. The Neighbor Decoder is the core component for length-insensitive STR, which takes $F^{(i)}$ as input and outputs string predictions $p^{(i)}$ and character features $g^{(i)}$. The Feature Enhancement Module enhances $F^{(i)}$ into $F^{(i+1)}$ with aid of $g^{(i)}$ with low computation cost. (b) The neighbor matrix generated by the Neighboring Navigator, which indicates the next neighbor of each character explicitly. (c) The feature map enhancer. The contextualized character features (in orange) are put back to $F^{(i)}$, then spread over $F^{(i)}$ by a convolution block.}
   \label{model_arch}
   \vspace{-2mm}
\end{figure*}


\section{Related Works}
We review previous arts on scene text recognition (STR) based on the aspects related to our contributions~\cite{zhu2016scene,long2021scene}.

\noindent \textbf{STR decoders.}
Text decoder is an essential component in a standard STR pipeline~\cite{Baek2019WhatIW}.
The CTC~\cite{Graves2006ConnectionistTC}-based decoder was adopted in various networks and applications~\cite{Shi2015crnn,Liu2016STARNetAS,Yousef2020OrigamiNetWS,Wang2021ImplicitFA,Hu2020GTCGT,Du2020PPOCRAP} due to its good balance between accuracy and efficiency. However, it was pointed out that the CTC loss misleads the feature alignments and representations\cite{Hu2020GTCGT}.
The serial attention mechanisms in STR~\cite{Shi2019ASTERAA,Lee2016RecursiveRN,Cheng2017FocusingAT,Li2018ShowAA,Cheng2021DecouplingVF,Li2021TrOCRTO,Lu2019MASTERMN,Tan2022PureTW,Bautista2022SceneTR} were inspired by the attention-based encoder-decoder framework in machine translation~\cite{Bahdanau2014NeuralMT} and the Transformer decoder~\cite{Vaswani2017AttentionIA}. Although the auto-regressive way achieved improved accuracy, the inference speed was also concerned since heavy computation remains in each decoding step.
As a result, the parallel attention mechanism has become popular in recent years~\cite{Wang2019DecoupledAN,Yu2020TowardsAS,Wan2019TextScannerRC,Fang2021ReadLH,Qiao2021PIMNetAP,Wang2021FromTT,Wang2022MultiGranularityPF,Na2022MultimodalTR}. It is efficient, but sensitive to text length for long text sequence~\cite{GarciaBordils2022OutofVocabularyCR}.
It is urgent to have a robust, effective and efficient decoder.

\noindent \textbf{Transformer in STR.} The Transformer~\cite{Vaswani2017AttentionIA} architecture has been widely used in STR.
Some works~\cite{Yu2020TowardsAS,Lu2019MASTERMN,Fang2021ReadLH} used convolution layers and Transformer encoder layers as the image feature encoder.
Recently, researchers built the backbone on the pure ViT~\cite{Dosovitskiy2020AnII,atienza2021vision,Wang2022MultiGranularityPF,Tan2022PureTW,Bautista2022SceneTR} and achieved impressive performance.
SVTR~\cite{Du2022SVTR} used stacked local and global mixing blocks that contain shift-window attention and standard self-attention layers.
Also, the Transformer decoder was directly taken as the text decoder in some arts~\cite{Lu2019MASTERMN,Tan2022PureTW,Bautista2022SceneTR}.
However, it is non-trivial to reduce the significant computation cost caused by the Transformer layers.
ABINet++~\cite{Fang2022ABINetAB} replaced the Transformer units with stacked convolution layers directly.
There are some works~\cite{Jiang2019VideoOD,Zhang2020KeyFP} that proposed to use key frame proposals for efficient video understanding.
Inspired by them, we only feed the key character features into the Transformer layers. Then the whole feature map is enhanced efficiently.

\noindent \textbf{Problem of text length in STR.} A key feature of the STR task is that text with arbitrary-length should be recognized.
However, most prevoius arts~\cite{Shi2015crnn,Shi2019ASTERAA,Yu2020TowardsAS,Fang2021ReadLH,Wang2022MultiGranularityPF} resized the input image to a fixed size and restricted the maximum length of text, which were unable to recognize unseen-length text absolutely.
Meantime, for text with rarely-seen length, serial and parallel attention-based mathods~\cite{Shi2019ASTERAA,Yu2020TowardsAS,Fang2021ReadLH,Tan2022PureTW,Na2022MultimodalTR} are both more prone to make errors~\cite{Wang2019DecoupledAN,GarciaBordils2022OutofVocabularyCR}.
Tan \etal ~\cite{Tan2022PureTW} designed an ensemble model with patches of different resolutions to adapt to different degrees of stretching in fixed-size images.
ABINet++~\cite{Fang2022ABINetAB} took as input images with a fixed bigger size, aggregated the horizontal features, and enriched the query vectors with content to alleviate the multi-activation phenomenon for long text.
In this work, we propose to build a text recognizer that is robust to text length.


\section{Methodology}
The proposed Length-Insensitive Scene Text Recognizer (LISTER) is able to read both short and long text images, and support iterative feature map enhancement.
As shown in \cref{model_arch}(a), LISTER consists of three parts: a feature extractor, a \textbf{Neighbor Decoder} (ND) and a \textbf{Feature Enhancement Module} (FEM).
A text image is first fed to the feature extractor, then the encoded feature map $F_{c\times h\times w}^{(0)}$ is passed to the length-robust ND to obtain the character attention maps$A^{(0)}$, align the character features $g^{(0)}$ and get predictions $p^{(0)}$ (decoding iter 0). We further enhance the feature map by capturing the long-range dependency of the aligned character features in FEM.
The enhanced feature map $F_{c\times h\times w}^{(i)}$ ($i>$0 means the feature map is enhanced $i$ times) is also used in ND (decoding iter 1, $\cdots$).

\subsection{Feature Extractor}
We adopt a convolution-based network, FocalNet~\cite{yang2022focal}, to extract visual features that does not acquire position encodings, which gets rid of the constraint of fixed image size. Similar to SVTR~\cite{Du2022SVTR}, we use strided depth-wise convolution and point-wise convolution to obtain image patch embeddings. Through the feature extractor, we obtain a feature map $F_{c\times h\times w}^{(0)}$ with rich local neighboring reception, where $c$ is the number of feature channels.


\subsection{Neighbor Decoder}
The attention-based neighbor decoder is able to translate feature maps into arbitrary-length text robustly.
Its attention mechanism consists of three simple modules, as shown in \cref{model_arch}(a). We first flatten $F_{c\times h\times w}^{(i)}(i\ge 0)$ as $\tilde{F}^{(i)}\in \mathbb{R}^{(hw)\times c}$.

\noindent \textbf{Neighboring Navigator} navigates each feature point in $\tilde{F}^{(i)}$ where its \textit{next neighbors} are. Here we demand that the next neighbors of a character-related point should be the points related to the next character.
To this end, the neighbor matrix $N$ is designed to indicate the next neighbors, as shown in \cref{model_arch}(b): $N_{jk}$ \textit{is the probability of the k-th point being the next neighbor of the j-th}.

Note that we append an [EOS] token to $\tilde{F}^{(i)}$, so that the last character has the special token as its neighbor. The resulted full feature sequence is:
\begin{equation}
   H^{(i)}=\left[\tilde{F}^{(i)};e_{\mathrm{EOS}}\right]\in \mathbb{R}^{S\times c}
   \label{cat_eos}
\end{equation}
where $[\cdot]$ is the concatenation operation, $e_{\mathrm{EOS}}$ is the learnable embedding of the [EOS] token, and $S=hw+1$. Then the neighbor matrix is obtained by a bilinear layer:
\begin{equation}
   N^{(i)}=\sigma\left(\frac{1}{\sqrt{c}}\left(H^{(i)}W_q\right)W_{r}\left(H^{(i)}W_k\right)^{\mathrm{T}} + b_r\right)
   \label{nb_attn}
\end{equation}
where $N^{(i)}\in \mathbb{R}^{S\times S}$, $W_q$, $W_k$ and $W_r$ are learnable weights with shape $c\times c$, $b_r\in \mathbb{R}$, and $\sigma$ is the softmax function.

\noindent \textbf{First Character Locator} generates the attention map of the first character. The query is a function of the global feature:
\begin{equation}
   q_0^{(i)}=\mathrm{GAP}(F^{(i)})W_q
\end{equation}
where $\mathrm{GAP}(\cdot)$ is the global average pooling. Then the attention map of the first character is:
\begin{equation}
   A_0^{(i)}=\sigma\left(\frac{q_0^{(i)}}{\sqrt{c}} \left(H^{(i)}W_k\right)\right)\in \mathbb{R}^{1\times S}
\end{equation}

\noindent \textbf{Character Aligner} produces all the character attention maps.
Ideally, if the ($j-1$)-th character has been located at the $J$-th feature in $H^{(i)}$, the next character can be located by looking up the neighbor matrix $N^{(i)}$ based on the index $J$, which is $\mathrm{argmax} N^{(i)}_J$.
In real case, we implement the idea in a soft way.
Given $N^{(i)}$ and $A_0^{(i)}$, $A^{(i)}_j (j>0)$ are calculated recurrently like a linked list:
\begin{equation}
   A^{(i)}_j = A^{(i)}_{j-1}N^{(i)}
   \label{attn_iter}
\end{equation}
The attention map generation is going on until
\begin{equation}
   A^{(i)}_{j,S-1}>\epsilon (0<\epsilon<1)
   \label{end_note}
\end{equation}
which means the end of decoding. $\epsilon$ is a hyper-parameter set to 0.6 empirically (the accuracy is insensitive to $\epsilon$).
Assuming that $L-1$ recurrences is performed, we obtain $A^{(i)}\in \mathbb{R}^{L\times S}$ which have $L-1$ real character attention maps and one [EOS] attention map.
Since $\sum _{k}A_{0,k}^{(i)}=1$, $\sum _{k}N^{(i)}_{lk}=1$, we can easily deduce that $\sum _{k}A_{j,k}^{(i)}=1$.
Hence, $A^{(i)}$ is a matrix of probability distributions, which meets the demand of been taken as the character attention maps here.

Although \cref{attn_iter} is performed in a serial way, it is still fast due to the low complexity of attention calculation. We will discuss it in the experiments.
In fact, \cref{attn_iter} is promising to run in parallel, since $A^{(i)}_j$ can also be calculated by only $A^{(i)}_0$ and $N^{(i)}$: $A^{(i)}_j = A^{(i)}_{0}{N^{(i)}}^j$.

The aligned character features are:
\begin{equation}
   g^{(i)}=A^{(i)}H^{(i)}\in \mathbb{R}^{L\times c}
\end{equation}
Finally, we get the text prediction by a linear layer followed by a softmax function.

\subsection{Feature Enhancement Module}
Long-range feature modeling is important in scene text recognition. The proposed Feature Enhancement Module (FEM) captures the long-range dependency of the aligned character features, and then utilizes them to enhance the previous feature map $F^{(i)}$ into $F^{(i+1)}$ with low computation cost. We do not use language embedding to be elegant.

Specifically, the character features $g^{(i)}$ are fed into a few transformer layers $\mathcal{T}$ with sliding windows~\cite{Beltagy2020LongformerTL}, to satisfy the original intention of processing arbitrary-length sequence:
\begin{equation}
   \tilde{g}^{(i)}=\mathcal{T}(g^{(i)})\in \mathbb{R}^{L\times c}
\end{equation}

\cref{model_arch}(c) illustrates the feature map enhancer.
The contextualized character features $\tilde{g}^{(i)}$ are \textit{put back} to the previous feature map, according to their attention maps:
\begin{equation}
   \tilde{G}^{(i+1)}=H^{(i)}+A^{(i)\mathrm{T}}\tilde{g}^{(i)}
\end{equation}
To further spread the long-range contextual information over the entire feature map, $\tilde{G}^{(i+1)}$ is reshaped to the size $h\times w \times c$ (the [EOS] feature is dropped), and a convolution block is simply exerted:
\begin{equation}
   F^{(i+1)}=\mathcal{C}\left(\tilde{G}^{(i+1)}\right)\in \mathbb{R}^{c\times h \times w}
\end{equation}


The enhanced feature map $F^{(i+1)}$ will also be decoded by the proposed neighbor decoder. The FEM+ND process can be performed iteratively to have finer and finer features.

\subsection{Training and Evaluation}
For each decoding iteration, the training objectives are three-fold as in \cref{eq_loss},
where $\lambda _1$ and $\lambda _2$ are set to $0.01$ and $0.001$ respectively.
The recognition loss $\mathcal{L}^{(i)}_{\mathrm{rec}}$ is the cross-entropy loss.
\begin{equation}
   \mathcal{L}^{(i)}=\mathcal{L}^{(i)}_{\mathrm{rec}}+\lambda _1 \mathcal{L}^{(i)}_{\mathrm{eos}}+\lambda _2 \mathcal{L}^{(i)}_{\mathrm{ent}}
   \label{eq_loss}
\end{equation}
To satisfy \cref{end_note} at the end of decoding, we design an ending location loss as in \cref{loss_eos},
where $A^{(i)}_{L-1}$ is the attention map for the [EOS] token.
\begin{equation}
   \mathcal{L}^{(i)}_{\mathrm{eos}}= - \mathrm{log}A^{(i)}_{L-1,S-1}
   \label{loss_eos}
\end{equation}
Besides, to avoid the multi-activation phenomenon~\cite{Fang2022ABINetAB}, the minimum entropy regularization is also used to make each attention map focus more on the core character region~\cite{cheng2020maximum}:
\begin{equation}
   \mathcal{L}^{(i)}_{\mathrm{ent}}=-\frac{1}{L}\frac{1}{\log (1+S)}\sum _j\sum _k A^{(i)}_{jk}\mathrm{log}A^{(i)}_{jk}
\end{equation}

For the model where FEM is iterated several times, the final objective takes the average loss.

During the inference, we propose an Attention Sharpening (AS) strategy to strengthen the ability for length extrapolation.
Noted that an imprecise attention map may not lead to an prediction error for its corresponding character directly, but could influence the latter ones by attention error accumulation.
Considering that an imprecise attention map usually has higher entropy, we propose to sharpen the character attention map when it is used for the next attention map generation. To this end, \cref{attn_iter} is re-writen to:
\begin{equation}
   A^{(i)}_j = \hat{A}^{(i)}_{j-1}N^{(i)}
\end{equation}
\begin{equation}
   \hat{A}^{(i)}_{j-1,s}=\frac{\exp\left(\alpha _j A^{(i)}_{j-1,s}\right)-1}{\sum _t \exp\left(\alpha _j A^{(i)}_{j-1,t}\right)-S}
\end{equation}
\begin{equation}
   \alpha _j=\min \left(1+\lambda (j-1), \mu\right)
\end{equation}
where $\lambda$ and $\mu$ are hyper-parameters set to 2 and 16 (insensitive) empirically. The resulted $\hat{A}^{(i)}_{j-1}$ is a sharpener distribution that could alleviate the accumulated attention errors.
Finally, the prediction of the last iteration is adopted.

\section{Experiments}

\subsection{Datasets} \label{sec:datasets}
\textbf{Common Benchmarks.} We simply use the common benchmark datasets (CoB) following ABINet~\cite{Fang2021ReadLH}, to make fair comparison. The two widely-used synthetic datasets, MJSynth (MJ)~\cite{Jaderberg2014SyntheticDA} and SynthText (ST)~\cite{Gupta2016SyntheticDF}, are used to train our model.
Six real datasets are usually used to evaluate STR models, including 3 regular datasets (IIIT5K~\cite{Mishra2009SceneTR} (3000), IC13~\cite{Karatzas2013ICDAR2R} (857) and SVT~\cite{Wang2011EndtoendST} (647)) and 3 irregular datasets (IC15~\cite{Karatzas2015ICDAR2C} (1811), SVTP~\cite{Phan2013RecognizingTW} (645) and CUTE~\cite{Risnumawan2014ARA} (288)).

\textbf{More challenging datasets.} Following PARSeq~\cite{Bautista2022SceneTR}, 3 challenging datasets are also taken for model evaluation: COCO-Text~\cite{Veit2016COCOTextDA} (9.8k), ArT~\cite{chng2019icdar2019} (35.1k) and Uber-Text~\cite{zhang2017uber} (80.6k).

Based on the length distribution of the synthetic training set, text of length longer than 16 is assumed as long here. Long text images account for 0.3\% in the 6 test sets of CoB, 1.1\% in ArT, 0.5\% in COCO-Text, and 1.9\% in Uber-Text.

\textbf{Text of Uniformly-distributed Lengths.} The existing popular benchmarks do not have enough long text images to evaluate length-insensitive text recognition well.
Hence, we collect a new scene text dataset as the new test set, named \textbf{Text of Uniformly-distributed Lengths} (TUL)\footnote{https://www.modelscope.cn/datasets/damo/TUL/summary}, where text of lengths 2-25 distributes uniformly, with 200 images and 200 different words for each length.
To be clear, we only consider 36 characters here, including 26 English letters and 10 digits.
The images are randomly sampled from the \textit{Out of Vocabulary Scene Text Understanding} competition dataset\footnote{https://rrc.cvc.uab.es/?ch=19\&com=introduction}~\cite{GarciaBordils2022OutofVocabularyCR}.
Images with very poor quality are filtered.
We suggest that models evaluated on TUL should not be trained on real training set~\cite{Bautista2022SceneTR,Baek2021WhatIW}, since there may be some overlaps between the real training data and TUL.

\subsection{Implementation Details}
For the feature extractor, the large receptive version of FocalNet~\cite{yang2022focal} is used except that the image width is only 4x downsampled.
For FEM, 8 heads are used in the self-attention layers with a window size of 11. \cref{ltr-variants} lists details of different LISTER variants. Throughout the experiments, we use the tiny version and 2 iterations of FEM (1 Transformer layer \& 1 Convolution block) by default, unless some specifications.
The number of classes is 37, including 26 lower-case letters, 10 digits and an end-of-sequence token.

\begin{table}[tp]
   \centering
   \caption{Architecture variants of LISTER.}
   \begin{tabular}{l|ccc}
      \toprule[1pt]
      \textbf{Model} & \textbf{Depths} & \textbf{Dimensions} & \textbf{FEM}\\
      \hline
      LISTER-T* & [2, 2, 6, 2] & [64, 128, 256, 512] & $\times$ \\
      LISTER-T & [2, 2, 6, 2] & [64, 128, 256, 512] & \checkmark \\
      LISTER-B* & [2, 2, 9, 2] & [96, 192, 384, 768] & $\times$ \\
      LISTER-B & [2, 2, 9, 2] & [96, 192, 384, 768] & \checkmark \\
      \bottomrule[1pt]
   \end{tabular}
   \label{ltr-variants}
   \vspace{-3mm}
\end{table}

\begin{table*}[htbp]
   \centering
   \caption{Comparison with other methods. The results of some methods on TUL are evaluated based on their publicly-released models. LISTER-T*$^\dagger$ is trained without $\mathcal{L}^{(i)}_{\mathrm{ent}}$. LISTER-B$^\#$ adopts the 3-scale ensemble strategy during inference.}
   \label{tab-sota}
   \begin{tabular}{l|ccc|ccc|c|c|c}
      \toprule[1pt]
      \multirow{2}*{\textbf{Method}} & \multicolumn{7}{c|}{\textbf{Common Benchmarks (CoB)}} & \multirow{2}*{\textbf{TUL}} & \multirow{2}*{\textbf{Params (M)}} \\
      \cline{2-8}
       & \textbf{IIIT5K} & \textbf{IC13} & \textbf{SVT} & \textbf{IC15} & \textbf{SVTP} & \textbf{CUTE} & \textbf{AVG} & & \\
      \hline
      SRN~\cite{Yu2020TowardsAS} & 94.8 & 95.5 & 91.5 & 82.7 & 85.1 & 87.8 & 90.4 & - & 54.7 \\
      ABINet$_{\textrm{vision}}$~\cite{Fang2021ReadLH} & 94.6 & 94.9 & 90.4 & 81.7 & 84.2 & 86.5 & 89.8 & 56.1 & 23.5\\
      VisionLAN~\cite{Wang2021FromTT} & 95.8 & 95.7 & 91.7 & 83.7 & 86.0 & 88.5 & 91.2 & - & 32.8 \\
      SVTR-L~\cite{Du2022SVTR} & 96.3 & 97.2 & 91.7 & 86.6 & 88.4 & \textbf{95.1} & 92.8 & - & 40.8 \\
      MGP-STR$_{\textrm{vision}}$~\cite{Wang2022MultiGranularityPF} & 96.4 & 96.5 & 93.2 & 86.3 & 89.5 & 90.6 & 92.7 & 50.6 & 85.5\\
      PARSeq$_N$~\cite{Bautista2022SceneTR} & 95.7 & 96.3 & 92.6 & 85.1 & 87.9 & 91.4 & 92.0 & - & 23.8\\
      PARSeq$_A$~\cite{Bautista2022SceneTR} & \textbf{97.0} & 97.0 & 93.6 & 86.5 & 88.9 & 92.2 & 93.2 & - & 23.8\\
      PTIE~\cite{Tan2022PureTW} & 96.3 & 97.2 & 94.9 & \textbf{87.8} & 90.1 & 91.7 & \textbf{93.5} & - & 45.9\\
      \hline
      ABINet~\cite{Fang2021ReadLH} & 96.2 & 97.4 & 93.5 & 86.0 & 89.3 & 89.2 & 92.6 & 57.7 & 36.7 \\
      MGP-STR~\cite{Wang2022MultiGranularityPF} & 96.4 & 97.3 & 94.7 & 87.2 & \textbf{91.0} & 90.3 & 93.3 & 50.8 & 148.0\\
      MATRN~\cite{Na2022MultimodalTR} & 96.6 & \textbf{97.9} & \textbf{95.0} & 86.6 & 90.6 & 93.5 & \textbf{93.5} & 60.5 & 44.2\\
      \hline
      LISTER-T*$^\dagger$ & 95.7 & 96.3 & 91.8 & 84.7 & 86.0 & 87.8 & 91.5 & 76.6 & 13.5 \\
      LISTER-T* & 96.0 & 96.3 & 92.7 & 84.9 & 86.4 & 85.1 & 91.7 & 76.8 & 13.5 \\
      LISTER-B* & 96.3 & 96.7 & 92.4 & 85.7 & 86.4 & 89.6 & 92.2 & 79.2 & 35.7 \\
      LISTER-T & 96.5 & 97.7 & 93.5 & 86.5 & 87.8 & 87.9 & 92.8 & 77.0 & 19.9 \\
      LISTER-B & 96.8 & 97.7 & 93.5 & 87.2 & 89.5 & 89.6 & 93.3 & 79.2 & 49.9 \\
      \hline
      LISTER-B$^\#$ & 96.9 & \textbf{97.9} & 93.8 & 87.5 & 89.6 & 90.6 & \textbf{93.5} & \textbf{79.5} & 49.9 \\
      \bottomrule[1pt]
   \end{tabular}
   \vspace{-4mm}
\end{table*}

\begin{table}
   \centering
   \caption{Comparison on more challenging datasets.}
   \label{tab-more_dataset}
   \begin{tabular}{l|ccc|c}
      \hline
      \textbf{Method} & \textbf{ArT} & \textbf{COCO} & \textbf{Uber} & \textbf{AVG} \\
      \hline
      ABINet~\cite{Fang2021ReadLH} & 65.4 & 57.1 & 34.9 & 45.2 \\
      MATRN~\cite{Na2022MultimodalTR} & 68.9 & 64.0 & 40.1 & 50.0 \\
      MGP-STR~\cite{Wang2022MultiGranularityPF} & 69.2 & 65.4 & 40.9 & 50.7 \\
      PARSeq$_N$~\cite{Bautista2022SceneTR} & 69.1 & 60.2 & 39.9 & 49.7\\
      PARSeq$_A$~\cite{Bautista2022SceneTR} & \textbf{70.7} & 64.0 & 42.0 & 51.8 \\
      \hline
      LISTER-T & 69.0 & 64.1 & 48.0 & 55.0 \\
      LISTER-B & 70.1 & \textbf{65.8} & \textbf{49.0} & \textbf{56.2} \\
      \hline
      LISTER-B$^h$ & \textbf{70.9} & \textbf{66.0} & \textbf{50.5} & \textbf{57.4} \\
      \hline
   \end{tabular}
   \vspace{-4mm}
\end{table}

For training, the model is initialized by using the truncated normal distribution.
The AdamW optimizer is used with weight decay of 0.05, and the initial learning rate is $1e-3$ which starts with 5000 warm-up steps and decays to $5e-7$ by a cosine scheduler.
The image augmentation technique in ABINet~\cite{Fang2021ReadLH} is used.
The height of images are set to 32, while the aspect ratios are randomly changed ranging from $\frac{1}{3}$ to 4 times.
In the batch training, the batch width is the maximum width of the batch samples. Shorter images are filled with zero pads.
Correspondingly, padding masks are exerted throughout the model.
To be efficient, we empirically restrict the maximum width to 256 and 416 for different settings.
There is no restriction on the maximum text length.
We train our model for 10 epochs with a batch size of 512 using one A100-80GB card.

For both training and evaluation, images with the widths $w<128$ are finally resized empirically\footnote{$h=32,w'=w\times 0.33+85$} to avoid the side effect of some narrow text images.
When comparing with PTIE~\cite{Tan2022PureTW} that proposed a strong ensemble way during evaluation, we also try a multi-scale ensemble strategy to improve the performance further. The input images are resized to 3 different scales\footnote{The other 2 scaling ops: $w'=w\times 0.26+95$, $w'=w\times 0.21+121$}, then fed to the model. The result with the highest probability is taken as the final prediction. It is very flexible and convenient since the input image size is unconstrained. we do not need to train an ensemble model.

\subsection{Comparison with State-of-the-arts}
We compare our LISTER with other methods on TUL and the common benchmarks respectively.

\noindent \textbf{Results on TUL.} To compare the performance for length-insensitive text recognition, we evaluate several representative methods that are strong on the common benchmarks recently, on the collected TUL.
As shown in \cref{tab-sota}, the total accuracies of them lag far behind the more lightweight LISTER-T* (improved by 16.3\% at least). The main reason lies in that they are poor for long text images, which is seen clearly in \cref{intro_len}.
As the model size increases and FEM is used, LISTER-B gets a further improvement of 2.5\%.

\noindent \textbf{Results on Common Benchmarks.} LISTER also does well on the common benchmarks where most texts are short.
As shwon in \cref{tab-sota}, LISTER-T* gets 91.7\% with only 13.5M parameters.
Enhanced by FEM, LISTER-T gets 92.8\% with 19.9 M parameters, reaching a good balance between accuracy and model size.
By scaling up, LISTER-B achieves an accuracy of 93.3\% that is nearly on par with the previous state-of-the-arts:
MGP-STR~\cite{Wang2022MultiGranularityPF} and MATRN~\cite{Na2022MultimodalTR} employed the external language priors to promote the final performance,
while PTIE used the multi-resolution patch ensemble strategy.
Through the simple 3-scale ensemble strategy (LISTER-B$^\#$), we obtain consistent gains, and achieves an average accuracy of 93.5\%.

\noindent \textbf{Results on Challenging Datasets.} As shown in \cref{tab-more_dataset}, LISTER-B outperforms others significantly on Uber-Text. One key factor may be the higher ratio of long text in Uber-Text.
However, LISTER-B does not get a better accuracy than PARSeq$_A$ on ArT where most cases have curved and rotated text.
It is mainly owed to the fact that the height of extracted feature map is squeezed to 1 in our implementation, which is unfriendly to irregular-shape text.
So we build another variant, LISTER-B$^h$, whose feature extractor outputs feature maps with $height=4$. As can be seen, LISTER-B$^h$ gets further improvement and catches the best.

\subsection{Impact of Neighbor Decoder}
The proposed neighbor decoder (ND) plays the key role for length-insensitive text recognition.
We conduct experiments on several popular decoders to compare their performance comprehensively: CTC, parallel attention (PAT), serial attention (SAT) and ND. The implementation of PAT is based on SRN~\cite{Yu2020TowardsAS}, and SAT is based on Aster~\cite{Shi2015crnn} that does not need additional pre-defined positional query embeddings.
To be fair, the experiments share all the settings except for the decoders, and SAT does not incorporate the internal language model.
The speed is measured on TUL by one Nvidia V100 GPU card.

\begin{figure}[t]
   \begin{center}
      \includegraphics[width=1.0\linewidth]{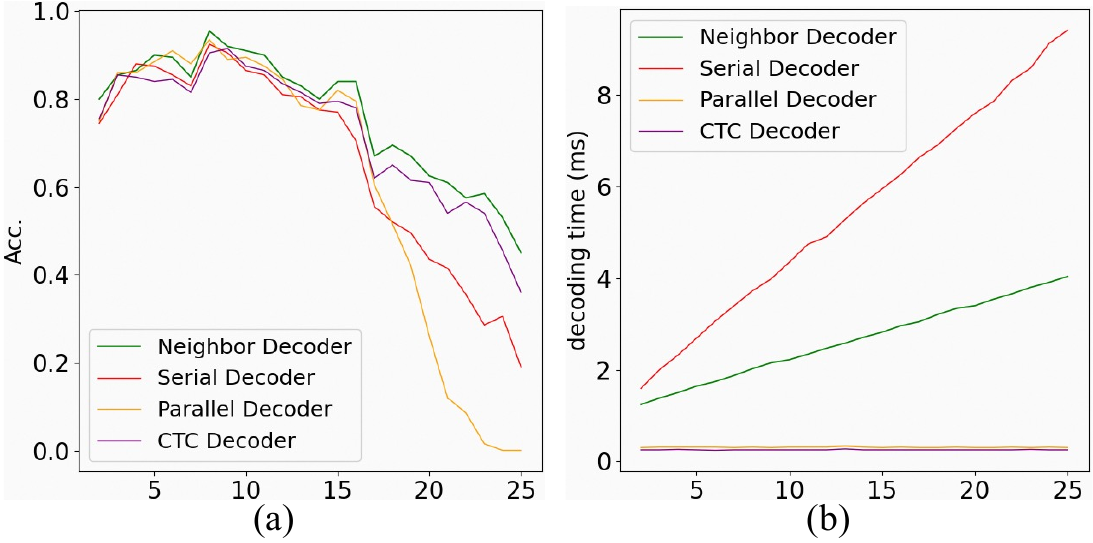}
   \end{center}
   \caption{Comparison of different decoders on TUL. (a) The accuracy distribution over different word lengths. (b) The decoding time distribution. (Best viewed in color.)}
   \label{fig_dec}
   \vspace{-2mm}
\end{figure}

\begin{table}
   \centering
   \caption{Results of different decoders.}
   \label{tab-dec}
   \begin{tabular}{l|c|c|m{.08\columnwidth}<{\centering}m{.08\columnwidth}<{\centering}m{.12\columnwidth}<{\centering}}
      \toprule[1pt]
      \textbf{Decoder} & \textbf{CoB} & \textbf{TUL} & \textbf{MACs (G)} & \textbf{Time (ms)} & \textbf{Params (M)}\\
      \hline
      CTC & 90.2 & 72.9 & \textbf{0.59} & \textbf{13.7} & \textbf{12.8} \\
      PAT & 91.4 & 61.6 & 0.60 & 13.8 & 13.0\\
      SAT & 91.3 & 66.5 & 0.73 & 18.1 & 15.4 \\
      \hline
      ND & \textbf{91.7} & \textbf{76.8} & 0.61 & 15.7 & 13.5\\
      \bottomrule[1pt]
   \end{tabular}
   \vspace{-2mm}
\end{table}

The results in \cref{tab-dec} and \cref{fig_dec} reveal that ND gets the best accuracies on both the common benchmarks and TUL with a little bit more cost.
PAT does well on the common benchmarks where short text accounts for the most, but gets much worse on TUL. The accuracy decreases dramatically as the text length increases beyond 16.
SAT has a slightly better accuracy on TUL with the cost of non-negligible MACs, latency and parameters. The attention-based RNN decoder still cannot adapt to long text well.
CTC is poor at the common short text recognition, but more robust to text lengths, which owes to the essence that it conducts the decoding by dense predictions and re-ordering the characters by a general rule~\cite{Graves2006ConnectionistTC}.

As for the decoding time, ND is slightly slower than CTC and PAT, but faster than SAT.
Assuming that the growth rate of the time complexity $O(N)$ is $\tau$, we find that $\tau$ of ND (0.122ms) is much less than that of SAT (0.340ms), as shown in \cref{fig_dec}(b). It is because the calculation in each attention generation step is very simple.

\begin{figure}[t]
   \begin{center}
      \includegraphics[width=1.0\linewidth]{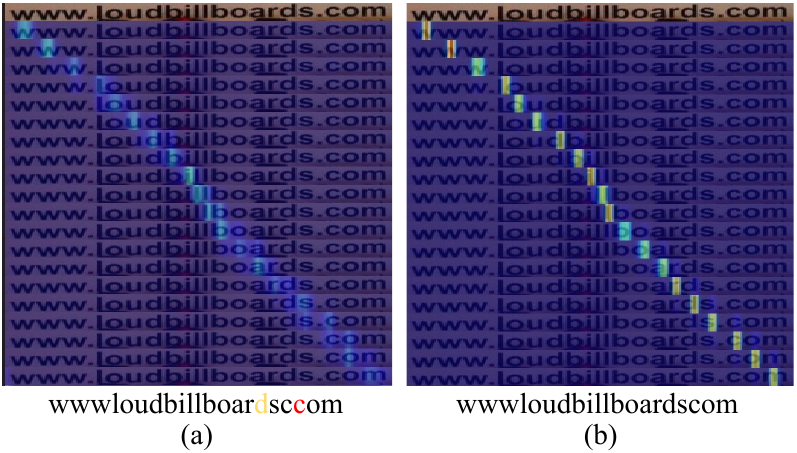}
   \end{center}
   \vspace{-3mm}
   \caption{Visualization of the attention maps for (a) parallel attention decoder and (b) the neighbor decoder. The character in yellow means it is missed, that in red means it is redundant or an error.}
   \label{exp_vis_dec}
   \vspace{-1mm}
\end{figure}

Character attention maps of a long text image are visualized in \cref{exp_vis_dec}. Obviously, the attention maps of the latter characters are vague for PAT, which leads to missing, redundant or error character predictions. For ND, the attention map is focusing and clear, which is promising to obtain the exact character features.

\subsection{Effects of Feature Enhancement Module}
The proposed FEM module is responsible for the long-range dependency modeling with low computation cost.
We study the effects of different iterations, the number of Transformer layers and convolution blocks, and where the Transformer layers should be placed.

\begin{table}
   \centering
   \caption{Ablation study on the FEM module. TrFE denotes that the Transformer layers are placed in the end of the feature extractor.}
   \label{tab-tfe}
   \begin{tabular}{m{.06\columnwidth}<{\centering}m{.05\columnwidth}<{\centering}m{.06\columnwidth}<{\centering}m{.07\columnwidth}<{\centering}|cc|m{.1\columnwidth}<{\centering}m{.1\columnwidth}<{\centering}}
      \toprule[1pt]
      \textbf{TrFE} & \textbf{No. Iters} & \textbf{No. Trans} & \textbf{No. Conv} & \textbf{CoB} & \textbf{TUL} & \textbf{Params (M)} & \textbf{MACs (G)}\\
      \hline
      \checkmark & - & - & - & 92.0 & 76.8 & 19.9 & 1.62\\
      \hline
       & 0 & - & - & 91.7 & 76.8 & 13.5 & 0.61\\
       & 1 & 2 & 2 & 92.6 & 77.2 & 26.2 & 1.03 \\
       & 2 & 0 & 1 & 92.0 & 76.5 & 16.7 & 0.85 \\
       & 2 & 1 & 0 & 92.2 & 76.4 & 16.7 & 0.76 \\
       & 2 & 1 & 1 & \textbf{92.8} & 77.0 & 19.9 & 1.05 \\
       & 2 & 2 & 2 & 92.7 & \textbf{77.8} & 26.2 & 1.45\\
      \bottomrule[1pt]
   \end{tabular}
   \vspace{-2mm}
\end{table}

From \cref{tab-tfe}, we find that increasing iterations of FEM leads to better performance with the increasing computation cost (1/2/2 VS 2/2/2).
The Transformer layer and the convolution block are both indispensable for FEM (by comparing 2/0/1, 2/1/0 and 2/1/1).
More Tansformer layers or convolution blocks contributes more on TUL (2/1/1 VS 2/2/2), which indicates that the long-range dependency in long text images is still important.

Some works~\cite{Yu2020TowardsAS,Fang2021ReadLH,Du2022SVTR} modeled the long-range dependency using Transformer layers in the latter part of their feature extractor. To explore the effects of them and ours fairly, we design a new setting: the Transformer layers are placed in the tail of the feature extractor too, without the FEM module.
The first line in \cref{tab-tfe} shows that it does not work as well as FEM, but forwards with more computation cost. It may be because that too much noisy background features are also modeled in the Transformer layers, which is lack of the focusing on the key character features.

\begin{figure}[t]
   \begin{center}
      \includegraphics[width=0.7\linewidth]{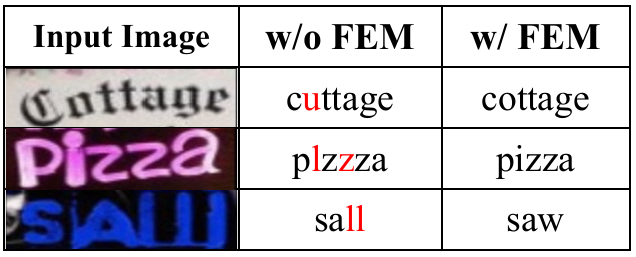}
   \end{center}
   \vspace{-2mm}
   \caption{Some recognition cases. The character in red means it is redundant or an error.}
   \label{fig-tfe}
   \vspace{-2mm}
\end{figure}

With iterations of FEM, the word prediction is refined. Surprisingly, FEM can edit the initial predicted wrong string by not only replacing characters, but also inserting and deleting.
\cref{fig-tfe} shows some cases.
This problem was pointed in SRN~\cite{Yu2020TowardsAS}, and explored by LevOCR~\cite{da2022levenshtein} interpretatively.

\subsection{Ability for Length Extrapolation}
An essential evidence for length-insensitive text recognition is the ability of length extrapolation, which means that the model should be able to recognize text of lengths unseen during training.
To verify it, we filter the synthetic training set by a simple rule: images with text lengths $> 16$ are dropped.
Then we train the models using the filtered training set. Note that we only train our LISTER and the CTC-based method for comparison. The PAT-based and SAT-based decoders are proven to perform poorly on long text images even under the training of full data in \cref{fig_dec}. The maximum width is increased to 416.

\begin{table}
   \centering
   \caption{The results of the ability for length extrapolation on TUL.}
   \label{tab_extra}
   \begin{tabular}{c|cc|ccc}
      \toprule[1pt]
      \textbf{Decoder} & \textbf{FEM} & \textbf{AS} & \textbf{Seen} & \textbf{Unseen} & \textbf{Total}\\
      \hline
      CTC & - & - & 82.9 & 56.2 & 72.9\\
      ND & - & - & 85.5 & 47.1 & 71.1\\
      ND & - & \checkmark & 85.8 & 58.4 & 75.5\\
      ND & \checkmark & \checkmark & \textbf{85.9} & \textbf{61.4} & \textbf{76.7}\\
      \bottomrule[1pt]
   \end{tabular}
   \vspace{-2mm}
\end{table}

\begin{figure}[t]
   \begin{center}
      \includegraphics[width=.8\linewidth]{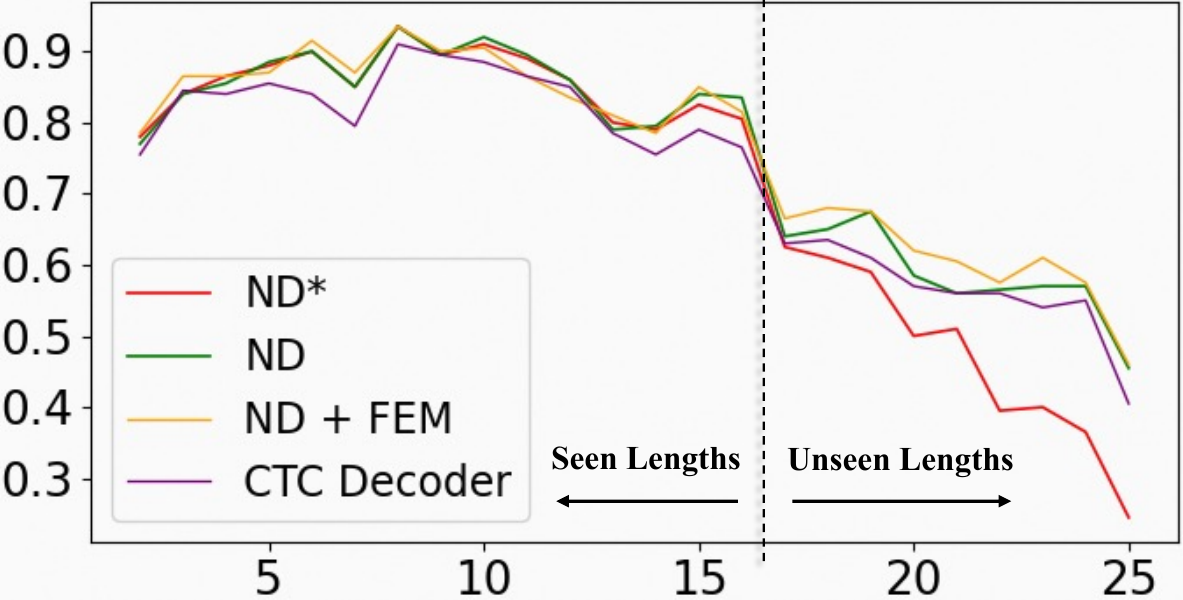}
   \end{center}
   \vspace{-3mm}
   \caption{Results about length extrapolation. ND* is the Neighbor Decoder version without attention sharpening.}
   \label{fig_extra}
   \vspace{-2mm}
\end{figure}

\begin{figure}[t]
   \begin{center}
      \includegraphics[width=1.0\linewidth]{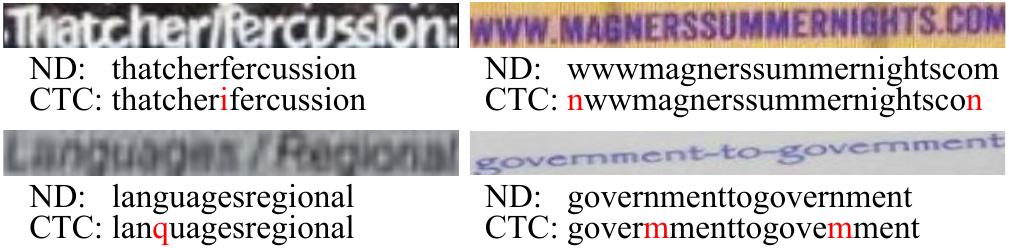}
   \end{center}
   \caption{Recognition cases of unseen lengths in the length extrapolation study. CTC is prone to make errors on similar-shape characters or be interrupted by special characters. Note that we do not use FEM on these cases to be fair.}
   \label{fig_case_extra}
   \vspace{-2mm}
\end{figure}

By the experiments, we conclude that ND have stronger ability for length extrapolation than CTC.
As shown in \cref{tab_extra} and \cref{fig_extra}, ND surpasses CTC on both seen- and unseen-length text. FEM brings a further improvement.
Some recognition cases are shown in \cref{fig_case_extra}.

\begin{figure}[t]
   \begin{center}
      \includegraphics[width=1.0\linewidth]{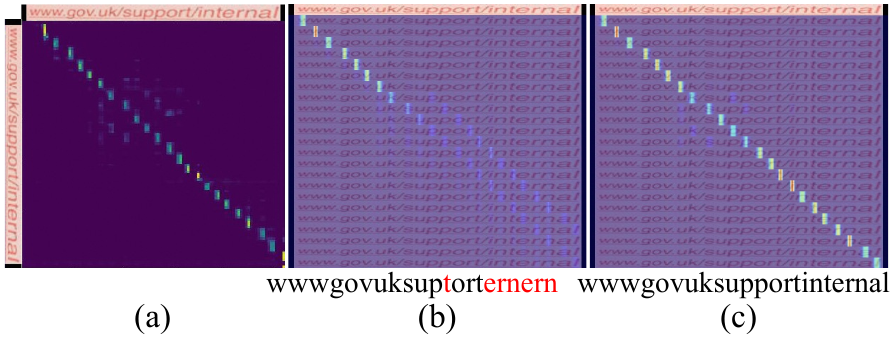}
   \end{center}
   \vspace{-3mm}
   \caption{Visualization of (a) the neighbor matrix, and attention maps of (b) ND without AS and (c) ND with AS, for an unseen-length sample.}
   \label{fig_vis_extra}
   \vspace{-4mm}
\end{figure}

A key factor for ND is the Attention Sharpening (AS) strategy. Without AS, ND does much worse than CTC on the unseen-length text.
By visualization as in \cref{fig_vis_extra}, we find that the neighbor matrix has no significant problems because the latter characters can still locate its next neighbor accurately.
However, the latter attention maps of ND without AS (\cref{fig_vis_extra}(b)) are vague, suffering a lot from the error accumulation. The little error in an early attention map will be enlarged as the attention map generation goes on.
After using AS, the attention map (\cref{fig_vis_extra}(c)) becomes clear and focusing, and the accuracy on the unseen-length text images increases by 11.3\% notably.

\subsection{Resizing or Padding?}
Most previous works adopted direct resizing to preprocess the images. Some arts pointed out that the input images should be resized rather than padding in their framework~\cite{Shi2019ASTERAA,Tan2022PureTW}.
However, direct resizing causes image distortion inevitably~\cite{Cheng2019PatchAF}, especially for the long text images.

For our framework, things have changed. As shown in \cref{tab_pad}, padding is always better for both PAT and ND.
It may owe to the local-aware feature extractor that does not model the global dependency frequently like RNN or the global self-attention layer, so our feature extractor is not that easy to overfitting on the length.

\begin{table}[t]
   \centering
   \caption{Comparison on the ways of image preprocessing.}
   \label{tab_pad}
   \begin{tabular}{c|cc|cc}
      \toprule[1pt]
      \textbf{Decoder} & \textbf{Padding} & \textbf{Resizing} & \textbf{CoB} & \textbf{TUL}\\
      \hline
      PAT & - & 128 & 91.2 & 55.4\\
      ND & - & 128 & 91.2 & 58.7\\
      ND & - & 256 & 90.9 & 65.9\\
      \hline
      PAT & \checkmark & - & 91.4 & 61.6\\
      ND & \checkmark & - & \textbf{91.7} & \textbf{76.8}\\
      \bottomrule[1pt]
   \end{tabular}
   \vspace{-2mm}
\end{table}

\subsection{Results on Chinese Text}
Instances of long text may frequently occur in some non-Latin scenarios, such as Chinese.
To demonstrate the superiority of ND in text images with real length distribution, we further conduct experiments on a Chinese scene text datase~\cite{Chen2021BenchmarkingCT} (scene-zh), where a highly imbalanced length distribution prevails (1-77).
The experiment settings follows Du \etal ~\cite{Du2022SVTR} to be fair, except that the maximum image width is extended to 448, and we do not restrict the maximum text length, which is an advantage of LISTER.

\begin{table}[t]
   \centering
   \caption{Comparison with other methods on scene-zh.}
   \label{tab_zh_sota}
   \begin{tabular}{c|c|c|c}
      \hline
      \textbf{SAR}~\cite{Li2018ShowAA} & \textbf{SRN}~\cite{Yu2020TowardsAS} & \textbf{SVTR-L}~\cite{Du2022SVTR} & \textbf{LISTER-B}\\
      \hline
      62.5 & 60.1 & 72.1 & \textbf{73.0}\\
      \hline
   \end{tabular}
   \vspace{-2mm}
\end{table}

\begin{table}[t]
   \centering
   \caption{Results of different decoders on scene-zh.}
   \label{tab_zh_dec}
   \begin{tabular}{c|m{.12\columnwidth}<{\centering}m{.12\columnwidth}<{\centering}m{.12\columnwidth}<{\centering}|m{.12\columnwidth}<{\centering}}
      \toprule[1pt]
      \textbf{Decoder} & \textbf{Short (59267)} & \textbf{Medium (3869)} & \textbf{Long (510)} & \textbf{Total (63646)}\\
      \hline
      PAT & 67.9 & 51.9 & 5.1 & 66.4\\
      CTC & 66.9 & 53.5 & 35.1 & 65.8\\
      \hline
      ND & \textbf{68.8} & \textbf{55.7} & \textbf{38.6} & \textbf{67.8}\\
      \bottomrule[1pt]
   \end{tabular}
   \vspace{-4mm}
\end{table}

Our LISTER-B achieves state-of-the-art accuracy, as shown in \cref{tab_zh_sota}.
We also compare our ND with PAT and CTC on the short, medium and long text respectively.
According to the length distribution, we manually regard texts of lengths $<12$ as short, that $>25$ as long, and others as medium.
From \cref{tab_zh_dec}, we conclude the same as discussed from \cref{tab-dec}, which confirms the strong ability of our length-insensitive text recognizer again, and the rationality of the collected TUL dataset.
From another perspective, LISTER is promising to adapt to multi-lingual text recognition.

\begin{figure}[t]
   \vspace{-2mm}
   \begin{center}
      \includegraphics[width=1.0\linewidth]{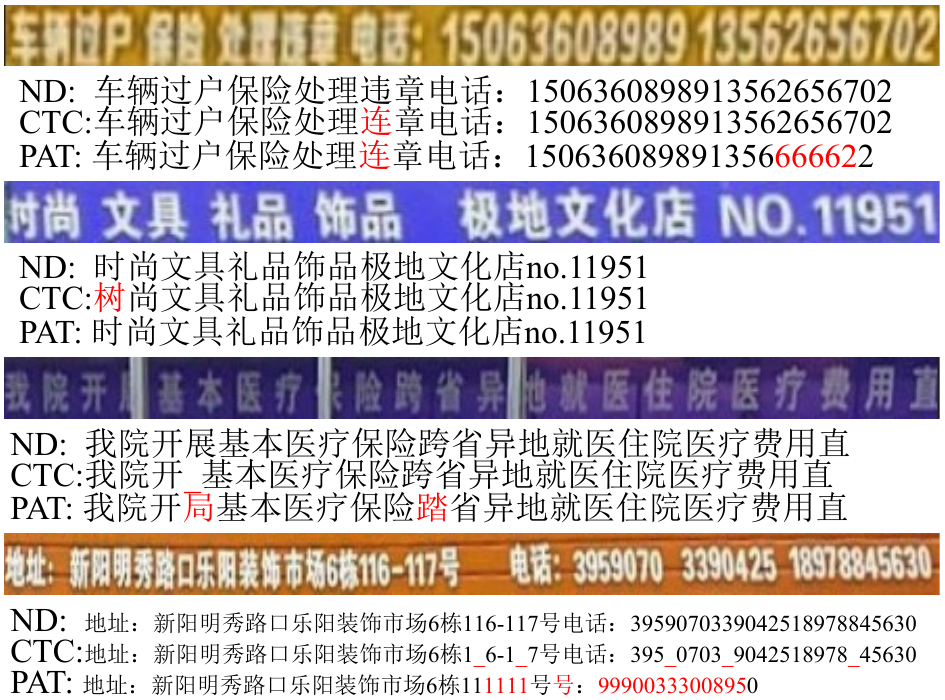}
   \end{center}
   \vspace{-2mm}
   \caption{Recognition cases in scene-zh. ``\textcolor{red}{\_}'' in red means the character is missed.}
   \label{fig_case_zh}
   \vspace{-3mm}
\end{figure}

Although CTC decoder is also robust to text length, it has two problems revealed in \cref{fig_case_zh}.
On the one hand, similar-shape characters are prone to be confused, as pointed in \cref{fig_case_extra} as well.
On the other hand, the congested successive same characters may be aggregated to only one due to the strict rule for character decoding (See the last case).
Our ND is both robust to text length as CTC, and effective on character feature extraction.

\section{Conclusion}
We have presented a length-insensitive scene text recognizer, LISTER, which is robust to text length.
As the core component, the attention-based neighbor decoder is able to obtain accurate attention maps through a novel neighbor matrix for text of arbitrary lengths robustly.
To model the long-range dependency with low computation cost, we propose a Feature Enhancement Module where only the aligned features are fed to the Transformer layers.
Extensive experiments have proved the effectiveness of LISTER on both short and long text images, as well as the capability of performing length extrapolation.
In the future, we will explore a more efficient and robust way of image feature encoding for length-insensitive STR.

{\small
\bibliographystyle{ieee_fullname}
\bibliography{egbib}

\begin{thebibliography}{10}\itemsep=-1pt

\bibitem{atienza2021vision}
Rowel Atienza.
\newblock Vision transformer for fast and efficient scene text recognition.
\newblock In {\em International Conference on Document Analysis and
  Recognition}, pages 319--334. Springer, 2021.

\bibitem{Baek2019WhatIW}
Jeonghun Baek, Geewook Kim, Junyeop Lee, Sungrae Park, Dongyoon Han, Sangdoo
  Yun, Seong~Joon Oh, and Hwalsuk Lee.
\newblock What is wrong with scene text recognition model comparisons? dataset
  and model analysis.
\newblock {\em 2019 IEEE/CVF International Conference on Computer Vision
  (ICCV)}, pages 4714--4722, 2019.

\bibitem{Baek2021WhatIW}
Jeonghun Baek, Yusuke Matsui, and Kiyoharu Aizawa.
\newblock What if we only use real datasets for scene text recognition? toward
  scene text recognition with fewer labels.
\newblock {\em 2021 IEEE/CVF Conference on Computer Vision and Pattern
  Recognition (CVPR)}, pages 3112--3121, 2021.

\bibitem{Bahdanau2014NeuralMT}
Dzmitry Bahdanau, Kyunghyun Cho, and Yoshua Bengio.
\newblock Neural machine translation by jointly learning to align and
  translate.
\newblock {\em CoRR}, abs/1409.0473, 2014.

\bibitem{Bautista2022SceneTR}
Darwin Bautista and Rowel Atienza.
\newblock Scene text recognition with permuted autoregressive sequence models.
\newblock {\em ArXiv}, abs/2207.06966, 2022.

\bibitem{Beltagy2020LongformerTL}
Iz Beltagy, Matthew~E. Peters, and Arman Cohan.
\newblock Longformer: The long-document transformer.
\newblock {\em ArXiv}, abs/2004.05150, 2020.

\bibitem{Chen2021BenchmarkingCT}
Jingye Chen, Haiyang Yu, Jianqi Ma, Mengnan Guan, Xixi Xu, Xiaocong Wang,
  Shaobo Qu, Bin Li, and X. Xue.
\newblock Benchmarking chinese text recognition: Datasets, baselines, and an
  empirical study.
\newblock {\em ArXiv}, abs/2112.15093, 2021.

\bibitem{Cheng2019PatchAF}
Changxu Cheng, Qiuhui Huang, Xiang Bai, Bin Feng, and Wenyu Liu.
\newblock Patch aggregator for scene text script identification.
\newblock {\em 2019 International Conference on Document Analysis and
  Recognition (ICDAR)}, pages 1077--1083, 2019.

\bibitem{Cheng2021DecouplingVF}
Changxu Cheng, Bohan Li, Qi Zheng, Yongpan Wang, and Wenyu Liu.
\newblock Decoupling visual-semantic feature learning for robust scene text
  recognition.
\newblock {\em ArXiv}, abs/2111.12351, 2021.

\bibitem{cheng2020maximum}
Changxu Cheng, Wuheng Xu, Xiang Bai, Bin Feng, and Wenyu Liu.
\newblock Maximum entropy regularization and chinese text recognition.
\newblock In {\em Document Analysis Systems: 14th IAPR International Workshop,
  DAS 2020, Wuhan, China, July 26--29, 2020, Proceedings 14}, pages 3--17.
  Springer, 2020.

\bibitem{Cheng2017FocusingAT}
Zhanzhan Cheng, Fan Bai, Yunlu Xu, Gang Zheng, Shiliang Pu, and Shuigeng Zhou.
\newblock Focusing attention: Towards accurate text recognition in natural
  images.
\newblock {\em 2017 IEEE International Conference on Computer Vision (ICCV)},
  pages 5086--5094, 2017.

\bibitem{chng2019icdar2019}
Chee~Kheng Chng, Yuliang Liu, Yipeng Sun, Chun~Chet Ng, Canjie Luo, Zihan Ni,
  ChuanMing Fang, Shuaitao Zhang, Junyu Han, Errui Ding, et~al.
\newblock Icdar2019 robust reading challenge on arbitrary-shaped text-rrc-art.
\newblock In {\em 2019 International Conference on Document Analysis and
  Recognition (ICDAR)}, pages 1571--1576. IEEE, 2019.

\bibitem{da2022levenshtein}
Cheng Da, Peng Wang, and Cong Yao.
\newblock Levenshtein {OCR}.
\newblock In {\em European Conference on Computer Vision}, pages 322--338.
  Springer, 2022.

\bibitem{Dosovitskiy2020AnII}
Alexey Dosovitskiy, Lucas Beyer, Alexander Kolesnikov, Dirk Weissenborn,
  Xiaohua Zhai, Thomas Unterthiner, Mostafa Dehghani, Matthias Minderer, Georg
  Heigold, Sylvain Gelly, Jakob Uszkoreit, and Neil Houlsby.
\newblock An image is worth 16x16 words: Transformers for image recognition at
  scale.
\newblock {\em ArXiv}, abs/2010.11929, 2020.

\bibitem{Du2022SVTR}
Yongkun Du, Zhineng Chen, Caiyan Jia, Xiaoyue Yin, Tianlun Zheng, Chenxia Li,
  Yuning Du, and Yu-Gang Jiang.
\newblock Svtr: Scene text recognition with a single visual model.
\newblock In {\em International Joint Conference on Artificial Intelligence},
  2022.

\bibitem{Du2020PPOCRAP}
Yuning Du, Chenxia Li, Ruoyu Guo, Xiaoting Yin, Weiwei Liu, Jun Zhou, Yifan
  Bai, Zilin Yu, Yehua Yang, Qingqing Dang, and Hongya Wang.
\newblock Pp-ocr: A practical ultra lightweight ocr system.
\newblock {\em ArXiv}, abs/2009.09941, 2020.

\bibitem{Fang2022ABINetAB}
Shancheng Fang, Zhendong Mao, Hongtao Xie, Yuxin Wang, Chenggang~Clarence Yan,
  and Yongdong Zhang.
\newblock Abinet++: Autonomous, bidirectional and iterative language modeling
  for scene text spotting.
\newblock {\em IEEE transactions on pattern analysis and machine intelligence},
  PP, 2022.

\bibitem{Fang2021ReadLH}
Shancheng Fang, Hongtao Xie, Yuxin Wang, Zhendong Mao, and Yongdong Zhang.
\newblock Read like humans: Autonomous, bidirectional and iterative language
  modeling for scene text recognition.
\newblock {\em 2021 IEEE/CVF Conference on Computer Vision and Pattern
  Recognition (CVPR)}, pages 7094--7103, 2021.

\bibitem{GarciaBordils2022OutofVocabularyCR}
Sergi Garcia-Bordils, Andr{\'e}s Mafla, Ali~Furkan Biten, Oren Nuriel, Aviad
  Aberdam, Shai Mazor, Ron Litman, and Dimosthenis Karatzas.
\newblock Out-of-vocabulary challenge report.
\newblock In {\em ECCV Workshops}, 2022.

\bibitem{Graves2006ConnectionistTC}
Alex Graves, Santiago Fern{\'a}ndez, Faustino~J. Gomez, and J{\"u}rgen
  Schmidhuber.
\newblock Connectionist temporal classification: labelling unsegmented sequence
  data with recurrent neural networks.
\newblock {\em Proceedings of the 23rd international conference on Machine
  learning}, 2006.

\bibitem{Gupta2016SyntheticDF}
Ankush Gupta, Andrea Vedaldi, and Andrew Zisserman.
\newblock Synthetic data for text localisation in natural images.
\newblock {\em 2016 IEEE Conference on Computer Vision and Pattern Recognition
  (CVPR)}, pages 2315--2324, 2016.

\bibitem{Hu2020GTCGT}
Wenyang Hu, Xiaocong Cai, Jun Hou, Shuai Yi, and Zhiping Lin.
\newblock Gtc: Guided training of ctc towards efficient and accurate scene text
  recognition.
\newblock In {\em AAAI Conference on Artificial Intelligence}, 2020.

\bibitem{Izmailov2018AveragingWL}
Pavel Izmailov, Dmitrii Podoprikhin, T. Garipov, Dmitry~P. Vetrov, and
  Andrew~Gordon Wilson.
\newblock Averaging weights leads to wider optima and better generalization.
\newblock In {\em Conference on Uncertainty in Artificial Intelligence}, 2018.

\bibitem{Jaderberg2014SyntheticDA}
Max Jaderberg, Karen Simonyan, Andrea Vedaldi, and Andrew Zisserman.
\newblock Synthetic data and artificial neural networks for natural scene text
  recognition.
\newblock {\em ArXiv}, abs/1406.2227, 2014.

\bibitem{Jiang2019VideoOD}
Zhengkai Jiang, Peng Gao, Chaoxu Guo, Qian Zhang, Shiming Xiang, and Chunhong
  Pan.
\newblock Video object detection with locally-weighted deformable neighbors.
\newblock In {\em AAAI Conference on Artificial Intelligence}, 2019.

\bibitem{Karatzas2015ICDAR2C}
Dimosthenis Karatzas, Llu{\'i}s~G{\'o}mez i Bigorda, Anguelos Nicolaou,
  Suman~K. Ghosh, Andrew~D. Bagdanov, M. Iwamura, Jiri Matas, Luk{\'a}s
  Neumann, Vijay~Ramaseshan Chandrasekhar, Shijian Lu, Faisal Shafait, Seiichi
  Uchida, and Ernest Valveny.
\newblock Icdar 2015 competition on robust reading.
\newblock {\em 2015 13th International Conference on Document Analysis and
  Recognition (ICDAR)}, pages 1156--1160, 2015.

\bibitem{Karatzas2013ICDAR2R}
Dimosthenis Karatzas, Faisal Shafait, Seiichi Uchida, M. Iwamura,
  Llu{\'i}s~G{\'o}mez i Bigorda, Sergi~Robles Mestre, Joan~Mas Romeu,
  David~Fern{\'a}ndez Mota, Jon Almaz{\'a}n, and Llu{\'i}s-Pere de~las Heras.
\newblock Icdar 2013 robust reading competition.
\newblock {\em 2013 12th International Conference on Document Analysis and
  Recognition}, pages 1484--1493, 2013.

\bibitem{Lee2016RecursiveRN}
Chen-Yu Lee and Simon Osindero.
\newblock Recursive recurrent nets with attention modeling for ocr in the wild.
\newblock {\em 2016 IEEE Conference on Computer Vision and Pattern Recognition
  (CVPR)}, pages 2231--2239, 2016.

\bibitem{Li2018ShowAA}
Hui Li, Peng Wang, Chunhua Shen, and Guyu Zhang.
\newblock Show, attend and read: A simple and strong baseline for irregular
  text recognition.
\newblock {\em ArXiv}, abs/1811.00751, 2018.

\bibitem{Li2021TrOCRTO}
Minghao Li, Tengchao Lv, Lei Cui, Yijuan Lu, Dinei A.~F. Flor{\^e}ncio, Cha
  Zhang, Zhoujun Li, and Furu Wei.
\newblock Trocr: Transformer-based optical character recognition with
  pre-trained models.
\newblock {\em ArXiv}, abs/2109.10282, 2021.

\bibitem{Liu2018ConnectionistTC}
Hu Liu, Sheng Jin, and Changshui Zhang.
\newblock Connectionist temporal classification with maximum entropy
  regularization.
\newblock In {\em Neural Information Processing Systems}, 2018.

\bibitem{Liu2016STARNetAS}
W. Liu, Chaofeng Chen, Kwan-Yee~Kenneth Wong, Zhizhong Su, and Junyu Han.
\newblock Star-net: A spatial attention residue network for scene text
  recognition.
\newblock In {\em British Machine Vision Conference}, 2016.

\bibitem{long2021scene}
Shangbang Long, Xin He, and Cong Yao.
\newblock Scene text detection and recognition: The deep learning era.
\newblock {\em International Journal of Computer Vision}, 129:161--184, 2021.

\bibitem{Lu2019MASTERMN}
Ning Lu, Wenwen Yu, Xianbiao Qi, Yihao Chen, Ping Gong, and Rong Xiao.
\newblock Master: Multi-aspect non-local network for scene text recognition.
\newblock {\em Pattern Recognit.}, 117:107980, 2019.

\bibitem{Lyu2018MaskTA}
Pengyuan Lyu, Minghui Liao, Cong Yao, Wenhao Wu, and Xiang Bai.
\newblock Mask textspotter: An end-to-end trainable neural network for spotting
  text with arbitrary shapes.
\newblock {\em IEEE Transactions on Pattern Analysis and Machine Intelligence},
  43:532--548, 2018.

\bibitem{Mishra2009SceneTR}
Anand Mishra, Alahari Karteek, and C.~V. Jawahar.
\newblock Scene text recognition using higher order language priors.
\newblock In {\em British Machine Vision Conference}, 2009.

\bibitem{Na2022MultimodalTR}
Byeonghu Na, Yoonsik Kim, and Sungrae Park.
\newblock Multi-modal text recognition networks: Interactive enhancements
  between visual and semantic features.
\newblock In {\em European Conference on Computer Vision}, 2022.

\bibitem{Phan2013RecognizingTW}
Trung~Quy Phan, Palaiahnakote Shivakumara, Shangxuan Tian, and Chew~Lim Tan.
\newblock Recognizing text with perspective distortion in natural scenes.
\newblock {\em 2013 IEEE International Conference on Computer Vision}, pages
  569--576, 2013.

\bibitem{Qiao2021PIMNetAP}
Zhi Qiao, Yu Zhou, Jin Wei, Wei Wang, Yuanqing Zhang, Ning Jiang, Hongbin Wang,
  and Weiping Wang.
\newblock Pimnet: A parallel, iterative and mimicking network for scene text
  recognition.
\newblock {\em Proceedings of the 29th ACM International Conference on
  Multimedia}, 2021.

\bibitem{Risnumawan2014ARA}
Anhar Risnumawan, Palaiahnakote Shivakumara, Chee~Seng Chan, and Chew~Lim Tan.
\newblock A robust arbitrary text detection system for natural scene images.
\newblock {\em Expert Syst. Appl.}, 41:8027--8048, 2014.

\bibitem{Shi2015crnn}
Baoguang Shi, Xiang Bai, and Cong Yao.
\newblock An end-to-end trainable neural network for image-based sequence
  recognition and its application to scene text recognition.
\newblock {\em IEEE Transactions on Pattern Analysis and Machine Intelligence},
  39:2298--2304, 2016.

\bibitem{Shi2016RobustST}
Baoguang Shi, Xinggang Wang, Pengyuan Lyu, Cong Yao, and Xiang Bai.
\newblock Robust scene text recognition with automatic rectification.
\newblock {\em 2016 IEEE Conference on Computer Vision and Pattern Recognition
  (CVPR)}, pages 4168--4176, 2016.

\bibitem{Shi2019ASTERAA}
Baoguang Shi, Mingkun Yang, Xinggang Wang, Pengyuan Lyu, Cong Yao, and Xiang
  Bai.
\newblock Aster: An attentional scene text recognizer with flexible
  rectification.
\newblock {\em IEEE Transactions on Pattern Analysis and Machine Intelligence},
  41:2035--2048, 2019.

\bibitem{Smith2017SuperConvergenceVF}
Leslie~N. Smith and Nicholay Topin.
\newblock Super-convergence: Very fast training of residual networks using
  large learning rates.
\newblock {\em ArXiv}, abs/1708.07120, 2017.

\bibitem{Tan2022PureTW}
Yew~Lee Tan, Adams Wai-Kin Kong, and Jung-Jae Kim.
\newblock Pure transformer with integrated experts for scene text recognition.
\newblock {\em ArXiv}, abs/2211.04963, 2022.

\bibitem{Vaswani2017AttentionIA}
Ashish Vaswani, Noam~M. Shazeer, Niki Parmar, Jakob Uszkoreit, Llion Jones,
  Aidan~N. Gomez, Lukasz Kaiser, and Illia Polosukhin.
\newblock Attention is all you need.
\newblock {\em ArXiv}, abs/1706.03762, 2017.

\bibitem{Veit2016COCOTextDA}
Andreas Veit, Tomas Matera, Luk{\'a}s Neumann, Jiri Matas, and Serge~J.
  Belongie.
\newblock Coco-text: Dataset and benchmark for text detection and recognition
  in natural images.
\newblock {\em ArXiv}, abs/1601.07140, 2016.

\bibitem{Wan2019TextScannerRC}
Zhaoyi Wan, Minghang He, Haoran Chen, Xiang Bai, and Cong Yao.
\newblock Textscanner: Reading characters in order for robust scene text
  recognition.
\newblock In {\em AAAI Conference on Artificial Intelligence}, 2019.

\bibitem{Wang2011EndtoendST}
Kai Wang, Boris Babenko, and Serge~J. Belongie.
\newblock End-to-end scene text recognition.
\newblock {\em 2011 International Conference on Computer Vision}, pages
  1457--1464, 2011.

\bibitem{Wang2022MultiGranularityPF}
Peng Wang, Cheng Da, and Cong Yao.
\newblock Multi-granularity prediction for scene text recognition.
\newblock In {\em European Conference on Computer Vision}, 2022.

\bibitem{Wang2019DecoupledAN}
Tianwei Wang, Yuanzhi Zhu, Lianwen Jin, Canjie Luo, Xiaoxue Chen, Y. Wu,
  Qianying Wang, and Mingxiang Cai.
\newblock Decoupled attention network for text recognition.
\newblock In {\em AAAI Conference on Artificial Intelligence}, 2019.

\bibitem{Wang2021ImplicitFA}
Tianwei Wang, Yuanzhi Zhu, Lianwen Jin, Dezhi Peng, Zhe Li, Mengchao He,
  Yongpan Wang, and Canjie Luo.
\newblock Implicit feature alignment: Learn to convert text recognizer to text
  spotter.
\newblock {\em 2021 IEEE/CVF Conference on Computer Vision and Pattern
  Recognition (CVPR)}, pages 5969--5978, 2021.

\bibitem{Wang2021FromTT}
Yuxin Wang, Hongtao Xie, Shancheng Fang, Jing Wang, Shenggao Zhu, and Yongdong
  Zhang.
\newblock From two to one: A new scene text recognizer with visual language
  modeling network.
\newblock {\em 2021 IEEE/CVF International Conference on Computer Vision
  (ICCV)}, pages 14174--14183, 2021.

\bibitem{yang2022focal}
Jianwei Yang, Chunyuan Li, Xiyang Dai, and Jianfeng Gao.
\newblock Focal modulation networks.
\newblock {\em Advances in Neural Information Processing Systems}, 2022.

\bibitem{Yousef2020OrigamiNetWS}
Mohamed Yousef and Tom~E. Bishop.
\newblock Origaminet: Weakly-supervised, segmentation-free, one-step, full page
  text recognition by learning to unfold.
\newblock {\em 2020 IEEE/CVF Conference on Computer Vision and Pattern
  Recognition (CVPR)}, pages 14698--14707, 2020.

\bibitem{Yu2020TowardsAS}
Deli Yu, Xuan Li, Chengquan Zhang, Junyu Han, Jingtuo Liu, and Errui Ding.
\newblock Towards accurate scene text recognition with semantic reasoning
  networks.
\newblock {\em 2020 IEEE/CVF Conference on Computer Vision and Pattern
  Recognition (CVPR)}, pages 12110--12119, 2020.

\bibitem{zhang2017uber}
Ying Zhang, Lionel Gueguen, Ilya Zharkov, Peter Zhang, Keith Seifert, and Ben
  Kadlec.
\newblock Uber-text: A large-scale dataset for optical character recognition
  from street-level imagery.
\newblock In {\em SUNw: Scene Understanding Workshop-CVPR}, volume 2017,
  page~5, 2017.

\bibitem{Zhang2020KeyFP}
Yuexi Zhang, Yin Wang, Octavia~I. Camps, and Mario Sznaier.
\newblock Key frame proposal network for efficient pose estimation in videos.
\newblock {\em ArXiv}, abs/2007.15217, 2020.

\bibitem{Zhong2022SGBANetSG}
Dajian Zhong, Shujing Lyu, Palaiahnakote Shivakumara, Bing Yin, Jiajia Wu,
  Umapada Pal, and Yue Lu.
\newblock Sgbanet: Semantic gan and balanced attention network for arbitrarily
  oriented scene text recognition.
\newblock In {\em European Conference on Computer Vision}, 2022.

\bibitem{Zhou2020InformerBE}
Haoyi Zhou, Shanghang Zhang, Jieqi Peng, Shuai Zhang, Jianxin Li, Hui Xiong,
  and Wan Zhang.
\newblock Informer: Beyond efficient transformer for long sequence time-series
  forecasting.
\newblock {\em ArXiv}, abs/2012.07436, 2020.

\bibitem{zhu2016scene}
Yingying Zhu, Cong Yao, and Xiang Bai.
\newblock Scene text detection and recognition: Recent advances and future
  trends.
\newblock {\em Frontiers of Computer Science}, 10:19--36, 2016.

\end{thebibliography}
}

\appendix
\section{Cases of Long Text Images}
\begin{figure}[t]
   \begin{center}
      \includegraphics[width=1.0\linewidth]{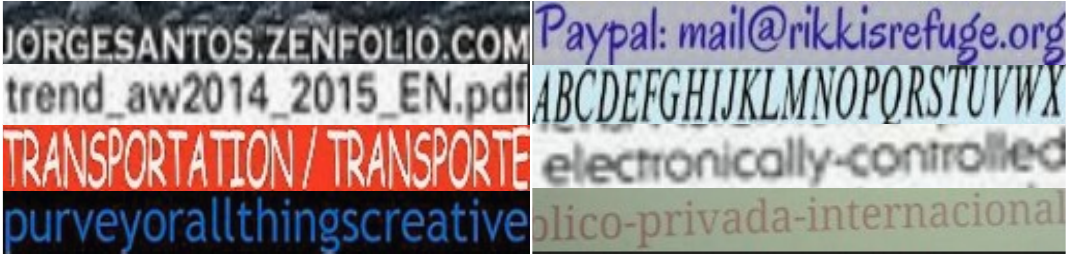}
   \end{center}
   \vspace{-2mm}
   \caption{Examples of long text in TUL.}
   \label{apdx_long}
\end{figure}

Long text occurs in our daily life frequently. It is a basic requirement to read long text for STR model.
\cref{apdx_long} gives some examples, including website, email address, file name, random code, compound word, etc.

\section{More Illustration on Attention Sharpening}
We would like to explain why we use Eq. (15) in the original paper to sharpen the character attention map.

Usually, the softmax function with temperature is exploited to sharpen probability distributions.
Suppose that we apply it to our attention sharpening directly,  \ie ,
\begin{equation}
   \hat{A}^{(i)}_{j-1,s}=\frac{\exp\left(\alpha _j A^{(i)}_{j-1,s}\right)}{\sum _t \exp\left(\alpha _j A^{(i)}_{j-1,t}\right)}
   \label{eq_softmax}
\end{equation}

Note that the exponential function can be converted as the following mathematically:
\begin{equation}
   e^x=1+x+o(x)
   \label{exp_exp}
\end{equation}
where $o(x)$ is a high-order infinitesimal. If $x_1$ and $x_2$ are two infinitesimals tending to 0, and if $x_1 < x_2$, then we have:
\begin{equation}
   \frac{e^{x_1}}{e^{x_1}+e^{x_2}}\approx \frac{1+x_1}{1+x_1 +1+x_2}>\frac{x_1}{x_1+x_2}
   \label{eq_cmp1}
\end{equation}
\begin{equation}
   \frac{e^{x_2}}{e^{x_1}+e^{x_2}}\approx \frac{1+x_2}{1+x_1 +1+x_2}<\frac{x_2}{x_1+x_2}
   \label{eq_cmp2}
\end{equation}
which means the discrepancy (normalized ratio) between $x_1$ and $x_2$ is reduced after the softmax.
In other words, the softmax function not only fails to sharpen the attention distribution, but flattens the distribution even more.
It is because $+1$ in \cref{exp_exp} dominates and dilutes $x$ in the normalization when $x$ is small.
Since $0\le A^{(i)}_{j-1,s}\le 1$, \cref{eq_softmax} has the same problem.

To avoid flattening the attention distribution, we simply \textit{replace the exponential function in softmax (\cref{eq_cmp1,eq_cmp2}) with} $e^x-1$.
In this way, \cref{eq_softmax} evolves into Eq. (15) in the original paper. In experiments, we find that Eq. (15) is more insensitive to $\alpha _j$ and achieves better results.

\section{Memory Access Cost}
The low memory access costs (MACs) is another advantage of the proposed LISTER, which allows us to use a large batch size.
As shown in \cref{tab_macs}, LISTER-B costs far less GPU memory than the hybrid convolution-Transformer architecture ABINet~\cite{Fang2021ReadLH} and the fully-Transformer network MGP~\cite{Wang2022MultiGranularityPF}.
We owe it to the depth-wise convolution in the feature extractor, the proposed simple neighbor decoder, and the sliding-window self-attention layer that only takes aligned character features as input in the proposed FEM. Besides, the height of the final feature map is 1, which also matters.

\begin{table}
   \centering
   \caption{Comparison on MACs. The methods are all tested with input of size $32\times 128$. For LISTER, the number of decoding steps is set to 12.}
   \label{tab_macs}
   \begin{tabular}{c|cc}
      \toprule[1pt]
      \textbf{Method} & \textbf{Params (M)} & \textbf{MACs (G)} \\
      \hline
      ABINet~\cite{Fang2021ReadLH} & 36.7 & 5.94\\
      MGP-STR$_{\textrm{vision}}$~\cite{Wang2022MultiGranularityPF} & 85.5 & 23.7\\
      \hline
      LISTER-B & 49.9 & 2.69\\
      \bottomrule[1pt]
   \end{tabular}
\end{table}

\section{Training using Real Dataset}
Recently, some works~\cite{Bautista2022SceneTR,Baek2021WhatIW} trained their models using real text dataset.
To evaluate LISTER more comprehensively, we further extend experiments by using the same real training dataset as in PARSeq~\cite{Bautista2022SceneTR}.
The 1cycle learning rate scheduler~\cite{Smith2017SuperConvergenceVF} and Stochastic Weight Averaging (SWA)~\cite{Izmailov2018AveragingWL} are also used during training.
The augmentation ways used in ABINet~\cite{Fang2021ReadLH} and PARSeq~\cite{Bautista2022SceneTR} are both exploited.
The maximum text length is restricted to 32 to be efficient during training, while arbitrary during inference.
The number of classes is still 37 to avoid inconsistence with the way of length calculation.

\subsection{Length Distribution Comparison}
\begin{figure}[t]
   \begin{center}
      \includegraphics[width=.99\linewidth]{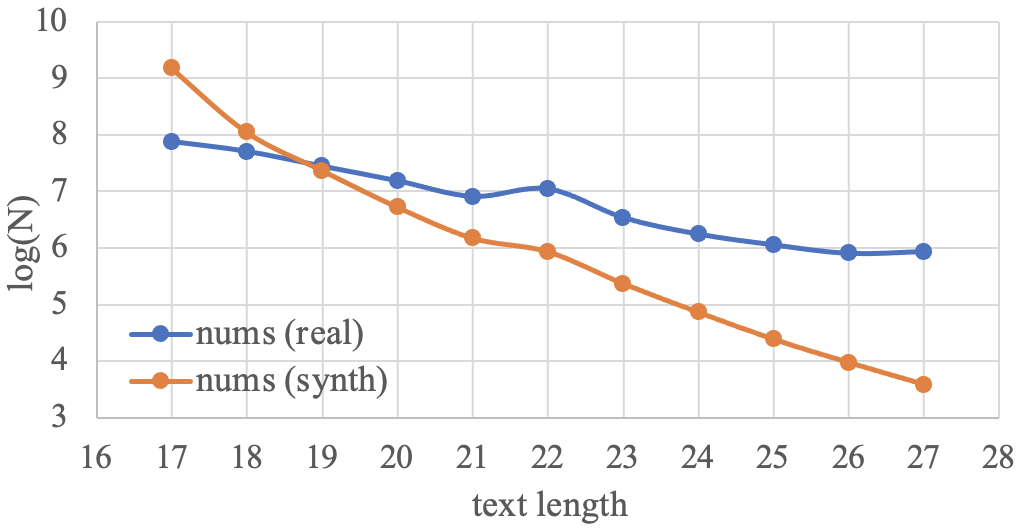}
   \end{center}
   \vspace{-2mm}
   \caption{Comparison on length distribution (17-27) between the real and synthetic training set. The number is in logarithmic.}
   \label{apdx_comp_distr}
\end{figure}

The real training set has much fewer samples than the widely-used synthetic dataset (MJ+ST), which is mainly reflected on short text.
However, the text length in the real has a wider distribution.
There are more long text images in the real set, as shown in \cref{apdx_comp_distr}.

\subsection{Results}
The results of models trained using real data are shown in \cref{apdx_comp_res}.
Among non-autoregressive models, LISTER performs the best except on ArT.
The gap between LISTER-B and LISTER-B$^h$ indicates that maintaining a proper height of feature map is necessary for irregular-shape text recognition.

The comparison on TUL is not appropriate here, since there are some overlaps between the real training set and TUL, as pointed in Sec. 4.1 in the paper. Nonetheless, LISTER achieves 88.6\% on TUL, which is convincing enough compared with PARSeq$_A$ (80.6\%).

\begin{table}
   \centering
   \caption{Comparison of models trained using real data. LISTER-B$^h$ is illustrated in Sec. 4.3. of the paper.}
   \label{apdx_comp_res}
   \begin{tabular}{l||c||ccc|c}
      \hline
      \textbf{Method} & \textbf{CoB} & \textbf{ArT} & \textbf{COCO} & \textbf{Uber} & \textbf{AVG} \\
      \hline
      \hline
      ABINet~\cite{Fang2021ReadLH} & 95.9 & 81.2 & 76.4 & 71.5 & 74.6 \\
      PARSeq$_N$~\cite{Bautista2022SceneTR} & 95.7 & \textbf{83.0} & 77.0 & 82.4 & 82.1 \\
      \hline
      LISTER-B & \textbf{96.4} & 81.8 & 77.0 & 79.4 & 79.9 \\
      LISTER-B$^h$ & 96.3 & 82.8 & \textbf{78.0} & \textbf{83.1} & \textbf{82.6} \\
      \hline
   \end{tabular}
   \vspace{-4mm}
\end{table}

\end{document}